\pdfoutput=1

\documentclass[11pt]{article}

\usepackage{EMNLP2023}
\usepackage{EMNLP2023}

\usepackage{times}
\usepackage{latexsym}
\usepackage{svg}
\usepackage{float}

\usepackage[utf8]{inputenc}
\usepackage{tikz}
\usetikzlibrary{trees}

\usepackage{amsmath}

\usepackage{tikz}
\usetikzlibrary{positioning,shapes.geometric}

\usepackage[T1]{fontenc}

\usepackage[utf8]{inputenc}

\usepackage{microtype}

\usepackage{inconsolata}
\usepackage{graphicx}
\usepackage{booktabs}
\usepackage{amssymb} 
\usepackage{tabularx} 

\usepackage{xcolor}
\usepackage[normalem]{ulem}

\def\MEdel#1{\bgroup\markoverwith{\textcolor{red}{\rule[0.5ex]{2pt}{1pt}}}\ULon{#1}}

\title{Merging in a Bottle: Differentiable Adaptive Merging (DAM) and the Path from Averaging to Automation}

\begin{document}



\author{
  Thomas Gauthier-Caron\textsuperscript{1},
  Shamane Siriwardhana\textsuperscript{1},
  Elliot Stein\textsuperscript{1},
  Malikeh Ehghaghi\textsuperscript{1},\\
  \bf Charles Goddard\textsuperscript{1},
  Mark McQuade\textsuperscript{1},
  Jacob Solawetz\textsuperscript{1},
  Maxime Labonne\textsuperscript{2} \\
  \\
  \textsuperscript{1}Arcee AI, San Francisco, USA \\
  \textsuperscript{2}Liquid AI, San Francisco, USA \\
  \\
  \texttt{\{thomas, shamane, elliot, malikeh, charles, mark,
  jacob\}@arcee.ai, maxime@liquid.ai} \\
}

\maketitle
\begin{abstract}

By merging models, AI systems can combine the distinct strengths of separate language models, achieving a balance between multiple capabilities without requiring substantial retraining. However, the integration process can be intricate due to differences in training methods and fine-tuning, typically necessitating specialized knowledge and repeated refinement. This paper explores model merging techniques across a spectrum of complexity, examining where automated methods like evolutionary strategies stand compared to hyperparameter-driven approaches such as DARE, TIES-Merging and simpler methods like Model Soups. In addition, we introduce Differentiable Adaptive Merging (DAM), an efficient, adaptive merging approach as an alternative to evolutionary merging that optimizes model integration through scaling coefficients, minimizing computational demands. Our findings reveal that even simple averaging methods, like Model Soups, perform competitively when model similarity is high, underscoring each technique’s unique strengths and limitations. We open-sourced DAM, including the implementation code and experiment pipeline, on GitHub\footnote{\href{https://github.com/arcee-ai/DAM}{https://github.com/arcee-ai/DAM}}.
\end{abstract}

\section{Introduction}
\label{sec:introduction}

As the demand for versatile and powerful AI systems grows, the need to merge Large Language Models (LLMs) with specialized capabilities, such as multilingual skills or domain-specific knowledge, has become increasingly pressing. Effective model merging enables systems to leverage the unique strengths of individual models without necessitating extensive retraining. Merging also offers the potential to reduce catastrophic forgetting, a significant advantage in maintaining learned knowledge from each model \cite{sukhbaatar2024branch,siriwardhana2024domain, labrak2024biomistral}. However, model merging remains inherently complex due to differences in training and fine-tuning processes, often requiring deep expertise and iterative tuning to achieve a balanced integration of the models' contributions.

Model merging techniques can be divided into two primary categories: manual and automated, and further distinguished by whether they are data-free or data-informed. Manual, data-free methods such as Model Soups \cite{wortsman2022model},  Trim, Elect, Sign, \& Merge (TIES-Merging) \cite{yadav2024ties} or Spherical Linear intERPolation (SLERP)\footnote{\href{https://github.com/Digitous/LLM-SLERP-Merge}{https://github.com/Digitous/LLM-SLERP-Merge}} focus on merging model parameters directly without any reliance on data, making them computationally efficient but requiring manual tuning, which can limit scalability. 

Automated, data-informed methods like AdaMerging \cite{yang2023adamerging} and evolutionary model merging \cite{akiba2024evolutionary} utilize representative data to inform and optimize parameter adjustments. This approach supports fine-grained control, such as per-layer or per-feature adjustments, reducing the need for manual tuning and improving performance on complex tasks. However, these automated methods typically demand more computational resources and may be impractical in scale. To gain deeper insight into the strengths and weaknesses of these approaches, we performed an in-depth comparative analysis of model merging techniques, spanning from basic averaging methods to more sophisticated automated approaches.



Building on these insights, we introduce Differentiable Adaptive Merging (DAM), a new approach developed as a more efficient alternative to compute-heavy evolutionary strategies. In this paper, we compare DAM with existing established techniques, including DRops And REscales with TIES sign election (DARE-TIES) \cite{yu2024language}, evolutionary merging \cite{akiba2024evolutionary}, and Model Soups \cite{wortsman2022model}, to provide a well-rounded view of model merging approaches. By positioning DAM as a cost-effective yet powerful alternative within this framework, we aim to highlight its practicality and effectiveness as a scalable merging solution.

Our contributions are as follows:

\begin{itemize}
    \item We conduct an extensive comparative analysis of model merging techniques, ranging from simple averaging methods to advanced automated approaches.
    \item We introduce DAM as a novel and efficient alternative to evolutionary merging, highlighting its effectiveness in reducing computational overhead while achieving competitive performance.
    \item Our findings underscore that simple methods, such as model averaging, can sometimes outperform more complex techniques, challenging established assumptions in the field and offering practical insights for researchers and practitioners. 
\end{itemize}

\section{Related Work}
\label{sec:related_work}

The process of model merging can broadly be split into pre-merge model alignment (if necessary), followed by a merging method which either uses the model weights alone, or some representative data samples to inform the merging process. We refer to these steps as \textbf{Model Alignment}, \textbf{Data-Free Merging} and \textbf{Data-Informed Merging} respectively.

\subsection{Model Alignment}

A fundamental issue in any model merging scenario is ensuring alignment. Alignment refers to the process of mapping functionally equivalent features to the same relative positions in the weight matrices of the models being merged. Without alignment, merging operations might combine disparate features, leading to interference and performance degradation.

Merging models trained from different initializations, or even on entirely different tasks, compounds this difficulty. Significant contributions to the field include methods like ZipIt \cite{stoica2023zipit}, Git Re-Basin \cite{ainsworth2022git}, and Optimal Transport (OT) Fusion \cite{singh2020model}. Git Re-Basin generalizes the merging process by permuting model parameters into a common space before averaging them, an approach that works well for models trained on similar tasks but struggles when tasks diverge significantly. OT Fusion employs Optimal Transport to achieve a similar goal, introducing soft matching for increased flexibility over strict permutations. ZipIt relaxes the constraints on merging, allowing for partial merging both across layers—merging up to a specified depth—and within layers by combining correlated features within models, as well as between models. This method excels in multi-task scenarios by accounting for unshared features, outperforming conventional permutation methods, especially when models are trained on disjoint tasks. Though even at their most successful, no current method has been shown to reliably merge a set of models trained from different initializations on different tasks and achieve high multi-task performance. This remains an open problem in the field.

Fortunately, when merging models fine-tuned from the same base or `parent' model, alignment is naturally preserved through the fine-tuning process. The merging methods discussed in subsequent sections assume this inherent alignment, enabling more effective integration of models.

\subsection{Data-Free Merging}

A significant subset of model merging methods focuses on combining model parameters through linear operations, often accompanied by pre-merging strategies to mitigate parameter interference. These methods consider only the model's parameters, making them highly computationally efficient compared to any LLM operation that involves data and inference. One of the pioneering methods in this space is Model Soups \cite{wortsman2022model}, which merges model weights via simple averaging. This approach first demonstrated the feasibility of weight-space merging for pre-trained LLMs and remains a reliable baseline due to its simplicity. However, despite its efficiency, the simplicity of Model Soup can lead to performance degradation caused by unresolved parameter conflicts.

A more sophisticated alternative to linear interpolation is SLERP\footnote{\href{https://github.com/Digitous/LLM-SLERP-Merge}{https://github.com/Digitous/LLM-SLERP-Merge}}, which interpolates between two models along the curved path connecting them on the surface of a sphere, with the base model serving as the center. SLERP is effective and widely adopted but limited by the fact that it can only merge two models at a time. Methods such as TIES-Merging \cite{yadav2024ties} and DELLA-Merging \cite{deep2024della} extend the basic principles of Model Soups by introducing refinements like trimming insignificant parameters, resolving sign conflicts, and pruning based on parameter magnitude. DARE \cite{yu2024language} further enhances this process by sparsifying delta parameters before merging, reducing redundancy and improving efficiency.

While these advanced techniques improve upon basic linear merging, they often require manual paramater tuning and are not fully automated. This lack of automation can hinder their scalability, particularly in complex multitask environments. Although their computational efficiency is a clear advantage, in scenarios where performance takes precedence over resource constraints, their inability to scale performance with available resources can be viewed as a limitation.

In practical applications, the open-source community frequently employs methods such as DARE-TIES \cite{yu2024language}, TIES \cite{yadav2024ties}, and SLERP. While these methods require careful hyperparameter tuning (for instance, to balance the weighting towards each input model), the community has developed general rules of thumb to guide this process. A commonly adopted approach, popularized by the MergeKit library \cite{goddard2024arcee}, involves assigning model parameters in a U-shaped pattern, with higher weights allocated to the initial and final layers, and lower weights assigned to the middle layers, relative to the base model. This method offers a balance between control and simplicity but still presents challenges in production environments. However, it remains somewhat unclear why these choices lead to effective results, as the underlying mechanics are not always fully understood. The black-box nature of these strategies highlights the need for more transparent, automated, and efficient model merging techniques.

\subsection{Data-Informed Merging}

Recognizing the limitations of manual parameter tuning, researchers have explored automated model merging techniques. These techniques leverage representative data during inference to better balance the trade-offs between the input models being merged. By utilizing the additional insights gained during inference, they facilitate the automatic selection of hyperparameters that would otherwise need to be approximated in data-free methods. Given these methods are fully automated, they can optimize more granular hyperparameters than the data-free methods, such as per-layer or even per-feature weighting parameters.


\textbf{AdaMerging} \cite{yang2023adamerging} is an advanced technique that automates model merging by focusing on minimizing the entropy associated with both the input models and their respective datasets. The process begins by calculating task vectors, which represent the difference in weights between the input models and the base model. Coefficients are then assigned and optimized for each task vector, determining the impact each input model has on the final merged model. AdaMerging adjusts these coefficients in an unsupervised manner, minimizing entropy on multi-task, unlabeled test data, which serves as a surrogate objective for minimizing loss on the full dataset. This optimization not only balances task vectors but also ensures the merged model’s predictions are more deterministic and robust. While AdaMerging has shown success in image classification tasks using transformer models, our work focuses on adapting this concept to merge LLMs for a broader range of tasks, including multilingual capabilities and domain-specific knowledge. 

\textbf{ZipLoRA} \cite{shah2023ziplora} is a technique developed to merge independently trained Low-Rank Adaptations (LoRAs) \cite{hu2021lora} by optimizing coefficients to reduce interference between content and style representations. They demonstrate that orthogonality between the columns of the LoRAs is strongly correlated with successful merging that minimizes interference. Primarily applied in the image generation domain with diffusion models, ZipLoRA preserves subject fidelity and style accuracy without requiring extensive manual adjustments. 

\textbf{Evolutionary Model Merging} \cite{akiba2024evolutionary} leverages evolutionary algorithms to optimize merging parameters across both the parameter space and the data flow (layer) space. Unlike AdaMerging, which assigns coefficients to task vectors, evolutionary techniques define merging coefficients on a per-layer, per-weight basis. The process involves initializing a population of potential solutions, each representing a different set of merging coefficients, and iteratively refining these through selection, mutation, and recombination based on task-specific performance metrics. Although evolutionary merging can produce highly effective merged models, it is computationally expensive. The need to evaluate numerous combinations of merging coefficients across large models with many layers demands substantial computational resources and time, making it less practical in scenarios with limited resources or time constraints.

\section{Differentiable Adaptive Merging (DAM)}
\label{sec:DAM}

In this section, we introduce DAM, a novel approach designed to merge multiple LLMs efficiently and effectively. DAM leverages a data-informed methodology to learn cost-efficient scaling coefficients for each model prior to merging, optimizing the integration process. This method can be applied to all linear layers, embedding layers, and layer normalization layers within the models.

\subsection{Mathematical Formulation}

The core idea of DAM is to find the optimal scaling coefficients for multi-task merging. By scaling the columns of the source models' weight matrices, we can effectively adjust the input features, ensuring that the merged model leverages the strengths of each individual model. For a given layer \( l \) in the model, let \( W_i^l \) represent the weight matrix of the \( i \)-th model. The goal is to find optimal scaling coefficients \( c_{ij}^l \) for each column \( j \) in the weight matrix of each model such that the merged weight matrix \( W^l \) is given by:

\[
W^l = \sum_{i=1}^{N} W_i^l \cdot C_i^l
\]

where \( N \) is the number of models being merged, and \( C_i^l \) is a diagonal matrix with the scaling coefficients \( c_{ij}^l \) on the diagonal. This ensures that each column of the weight matrix is scaled individually.

\subsection{Column-wise Scaling}

\begin{figure*}[htbp]
  \centering
  \includegraphics[scale=0.4]{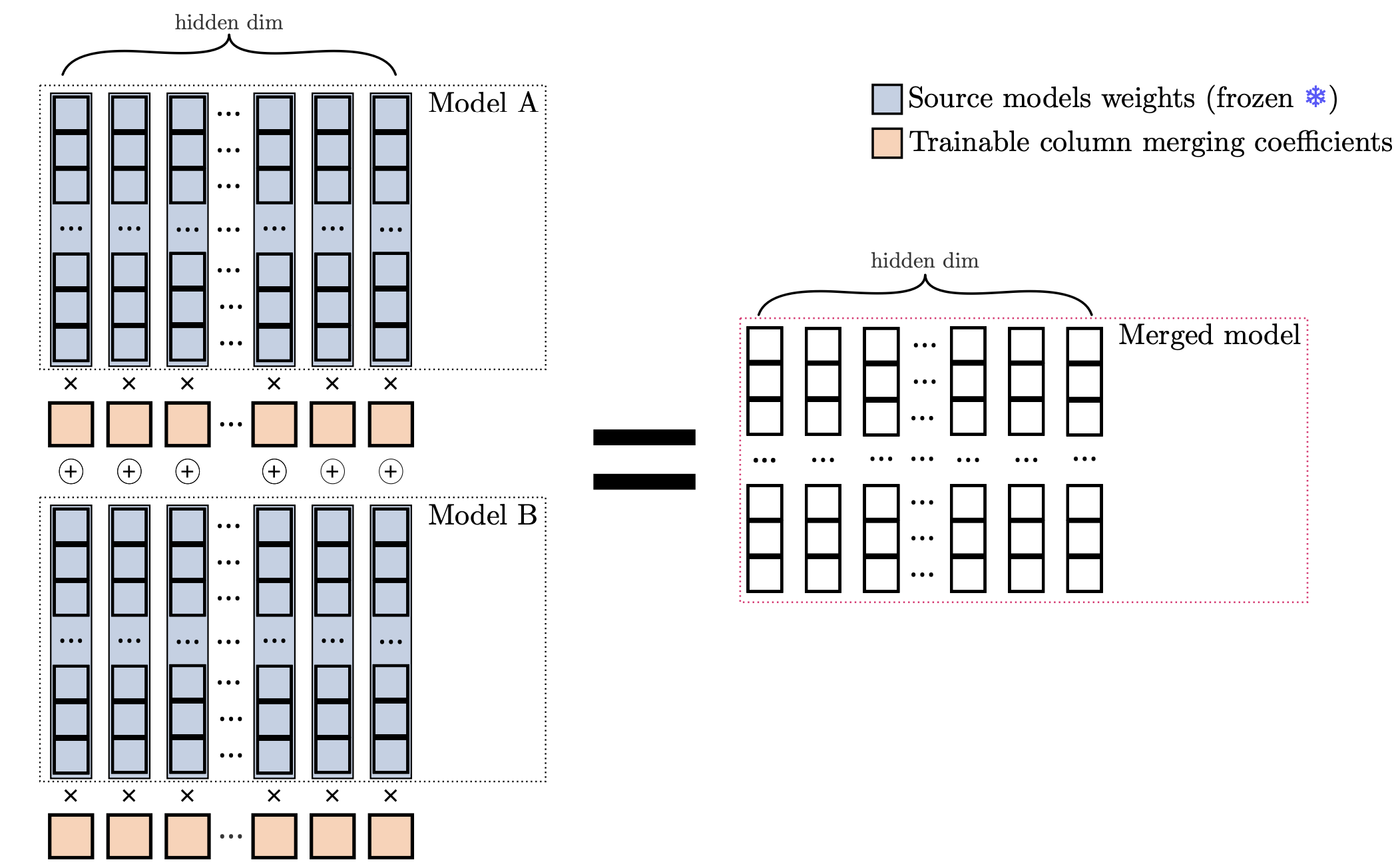}
  \caption{Graphical illustration of activation adjusting via the trainable coefficients in the proposed DAM method.}
  \label{fig:dam_activation_coefficients}
\end{figure*}

To provide a more intuitive understanding, consider a weight matrix \( W_i^l \) of layer \( l \) in model \( i \) with dimensions \( M \times N \). Each column \( j \) in this weight matrix has a corresponding scaling factor \( c_{ij}^l \). The merged weight matrix \( W^l \) can then be expressed as:

\[
W^l = \sum_{i=1}^{N} W_i^l \cdot \text{diag}(c_{i1}^l, c_{i2}^l, \ldots, c_{iN}^l)
\]

where \( \text{diag}(c_{i1}^l, c_{i2}^l, \ldots, c_{iN}^l) \) is a diagonal matrix with the scaling factors on the diagonal. These factors are learned to optimally scale the input features, effectively adjusting the contribution of each model's weight matrix to the merged model. See Figure~\ref{fig:dam_activation_coefficients}.

For example, consider the MLP (Multi-Layer Perceptron) layers in three models. Each model has its own weight matrix for a given layer. By learning the optimal scaling factors for each column of these weight matrices, we can merge the models in a way that scales the input features appropriately, ensuring that the merged model performs optimally across tasks.

The DAM method is applied uniformly across different types of layers in the models: for each linear layer, the weight matrices \( W_i^l \) are scaled and merged using the learned coefficients \( c_{ij}^l \); similarly, the embedding matrices in the embedding layers are scaled and merged; and in the layer normalization layers, the normalization parameters are scaled and merged to ensure consistent normalization across the merged model. In our experiments, we found that merging only the linear layers, while retaining the embeddings and normalization weights from the base model, performs best.

\subsection{Objective Function for DAM}\label{Objective  Function section}

The objective function for DAM is designed to optimize the scaling coefficients for each model in a way that ensures the best possible merge. The core idea is to balance task performance, regularization, and similarity constraints between coefficients to achieve a robust and efficient merging process. The objective function consists of the following components:

\subsubsection{Kullback-Leibler (KL) Divergence Loss}

Given three models and their corresponding datasets \( D_1, D_2, D_3, .. , D_n \), the idea is to minimize the KL divergence between the logits of the merged model and the logits of each individual model on their respective datasets. Let \( \text{KL}(P \| Q) \) denote the KL divergence between distributions \( P \) and \( Q \). The KL divergence loss for the merged model is given by:

\[
\mathcal{L}_{\text{KL}} = \sum_{i=1}^{N} \text{KL}(\text{logits}_{\text{merged}}(D_i) \| \text{logits}_{i}(D_i))
\] \label{KL Div Loss}

\noindent where \( \text{logits}_{\text{merged}}(D_i) \) are the logits of the merged model on dataset \( D_i \), and \( \text{logits}_{i}(D_i) \) are the logits of the \( i \)-th model on its corresponding dataset. This ensures that the merged model's predictions are aligned with those of the individual models on their respective datasets.






\subsubsection{Cosine Similarity Loss}

Inspired by ZipLoRA, we add a constraint to reduce the cosine similarity between the scaling coefficients of different models for each layer. This encourages the models to scale the feature space in unique ways, promoting diversity in the merged model. The cosine similarity loss is given by:

\[
\mathcal{L}_{\text{cosine}} = \lambda_{\text{cosine}} \sum_{l} \sum_{i < j} \text{cos}(c_{ij}^l, c_{ik}^l)
\]

where \( \lambda_{\text{cosine}} \) is the regularization coefficient for the cosine similarity loss. This term helps to ensure that the scaling coefficients for different models are not too similar, which can improve the robustness of the merged model.

\subsubsection{L1 and L2 Regularization}

To ensure stable training and add sparsity to the coefficients, we apply L1 and L2 regularization to the scaling coefficients:

\[
\mathcal{L}_{\text{reg}} = \lambda_1 \sum_{i=1}^{N} \sum_{j=1}^{d} \|c_{ij}^l\|_1 + \lambda_2 \sum_{i=1}^{N} \sum_{j=1}^{d} \|c_{ij}^l\|_2^2
\]

where \( \lambda_1 \) and \( \lambda_2 \) are regularization coefficients. L1 regularization encourages sparsity in the scaling coefficients, while L2 regularization ensures that the coefficients remain small and stable during training.





\subsubsection{Overall Objective Function}

The overall objective function for training the scaling coefficients in DAM combines the KL divergence loss, cosine similarity loss, and regularization terms:

\[
\mathcal{L} = \mathcal{L}_{\text{KL}} + \mathcal{L}_{\text{cosine}} + \mathcal{L}_{\text{reg}}
\]\label{objective function}

This comprehensive objective function ensures that the scaling coefficients are optimized for task performance, regularization, and similarity, leading to a more robust and efficient merging process.

\section{Experiment Design}


\label{sec:experiments}
\subsection{Benchmarking Different Merging Techniques against DAM}
To thoroughly benchmark model merging approaches, we designed experiments focusing on two primary model families, Mistral \cite{jiang2023mistral} and Llama 3 \cite{dubey2024llama}. In these experiments, we compare the effectiveness of various merging methods across models with diverse specialized capabilities, including multilingual processing, coding proficiency, and mathematical reasoning. This approach allows us to evaluate how well each merging technique retains and integrates these distinct functionalities within the merged models.

Given the proprietary nature of the original datasets used to train each of the candidate models, we applied closely aligned, representative datasets specifically tailored for DAM training. This approach ensures that the DAM process reflects the specialized capabilities of each model.





\subsubsection{Case Study 1: Japanese Language Processing and Mathematical Reasoning}
In this case study, we applied Mistral-based models specialized in Japanese language processing and mathematical reasoning to evaluate the effectiveness of various merging techniques in preserving and integrating these distinct capabilities. Our experiments focus on assessing the adaptability and compositional generalization potential of DAM by examining its performance on mathematical reasoning tasks in Japanese. To benchmark DAM’s capabilities, we compared it against other merging techniques, including DARE-TIES, Model Soups, and evolutionary merging methods.



\textbf{Selected Models: }For this case study, we selected the \texttt{shisa-gamma-7b} model \cite{shisagamma7b}, a Japanese language model trained specifically on Japanese language tasks, demonstrating proficiency in linguistic comprehension and expression in Japanese. Additionally, we utilized the \texttt{WizardMath-7B-V1.1} model \cite{luo2023wizardmathempoweringmathematicalreasoning} and \texttt{Abel-7B-002} model \cite{abel}, both of which are trained on datasets oriented towards mathematical reasoning, equipping them with capabilities in numerical and logical problem-solving. Each of these models is derived from the \texttt{Mistral-7B-v0.1} base model \cite{mistral}, and our merged models retain the embedding and RMSNorm weights from \texttt{Mistral-7B-v0.1} to maintain consistency.

\textbf{Representative Datasets: } For DAM training, we used the \texttt{Ichikara} Japanese instruction tuning dataset \cite{sekine2023ichikara}, covering the same kind of broad general-purpose Japanese language chat conversations used to train \texttt{shisa-gamma-7b}. For mathematical reasoning, we used the \texttt{MetaMathQA} \cite{yu2023metamath} and \texttt{Orca-Math} \cite{mitra2024orcamath} datasets, covering basic arithmetic, algebraic operations, and logical reasoning to simulate the reasoning tasks likely included in \texttt{WizardMath} and \texttt{Abel}. The final training dataset comprises 1,729 samples, formatted with the Alpaca instruction template \cite{taori2023alpaca} to ensure consistency. DAM training was conducted with a learning rate of \(2e^{-3}\) and a batch size of 1 to balance linguistic and mathematical capabilities across the merged models.


\textbf{Evaluation Metrics: }To assess the performance of the merged models in Japanese and math, we used the JP Language Model Evaluation Harness\footnote{\href{https://github.com/Stability-AI/lm-evaluation-harness/tree/jp-stable}{https://github.com/Stability-AI/lm-evaluation-harness/tree/jp-stable}}. We evaluated Japanese language proficiency of the merged models through a range of metrics for linguistic accuracy and comprehension in Japanese, as shown in Table \ref{table:method-comparison}, and we tested mathematical reasoning abilities on the Japanese subset of the Multilingual Grade School Math (MGSM) Benchmark \cite{shi2022language}. 


\subsubsection{Case Study 2: SQL Coding, German, and Korean Language Processing}

In this case study, we expanded our experiments to encompass multilingual and domain-specific models in German, Korean, and SQL, aiming to further evaluate DAM’s effectiveness in managing diverse languages and tasks. In this context, we conducted comparative evaluations of DARE-TIES, Model Soups, against DAM, assessing their respective performances across these varied domains. 



\textbf{Selected Models: }For this case study, we used the following Llama-3-based models: \texttt{Llama-3-SauerkrautLM-8b-Instruct}\footnote{\href{https://huggingface.co/VAGOsolutions/Llama-3-SauerkrautLM-8b-Instruct}{https://huggingface.co/VAGOsolutions/Llama-3-SauerkrautLM-8b-Instruct}}, a German language model fine-tuned for various linguistic tasks in German; \texttt{Llama-3-Open-Ko-8B}\footnote{\href{https://huggingface.co/beomi/Llama-3-Open-Ko-8B}{https://huggingface.co/beomi/Llama-3-Open-Ko-8B}}, specialized in Korean language processing; and \texttt{llama-3-sqlcoder-8b}\footnote{\href{https://huggingface.co/defog/llama-3-sqlcoder-8b}{https://huggingface.co/defog/llama-3-sqlcoder-8b}}, tailored for SQL coding tasks. Each of these models is based on the \texttt{Meta-Llama-3-8B}\footnote{\href{https://huggingface.co/meta-llama/Meta-Llama-3-8B}{https://huggingface.co/meta-llama/Meta-Llama-3-8B}} model, with the German and SQL models being initialized from the Instruct variant.

\textbf{Representative Datasets: } For German, we sourced diverse linguistic samples from corpora, using \texttt{Fischerboot/another-german-alpaca-dataset}\footnote{\url{https://huggingface.co/datasets/Fischerboot/another-german-alpaca-dataset}}, covering comprehension and dialogue tasks. For Korean, we included samples in both formal and informal registers from \texttt{lcw99/wikipedia-korean-20240501}\footnote{\url{https://huggingface.co/datasets/lcw99/wikipedia-korean-20240501}} for broader contextual understanding. SQL query examples were used from the \texttt{wikiSQL} \cite{zhongSeq2SQL2017}, emphasizing syntactical accuracy and command structure. All tasks were formatted using the Alpaca instruction template to maintain consistency \cite{taori2023alpaca}. Training parameters mirrored our prior study, with a learning rate of $(2e^{-3})$ and a batch size of 1, achieving balanced performance across multilingual and domain-specific models.

\textbf{Evaluation Metrics: } To evaluate the language capabilities of the merged models in German, we used the German subset of the Okapi benchmark \cite{lai2023okapi}, which provides a comprehensive set of tasks for assessing model's language comprehension. For Korean, we employed the Korean Balanced Evaluation of Significant Tasks (KoBEST) benchmark \cite{jang2022kobest}, which consists of five distinct Korean-language tasks designed to gauge various aspects of language comprehension in Korean, including factual reasoning, causality, word sense disambiguation, contextual inference, and sentiment analysis. For SQL generation, we measured model performance using the SQL-Eval\footnote{\href{https://github.com/defog-ai/sql-eval}{https://github.com/defog-ai/sql-eval}} framework, focusing on the syntactical correctness and structural accuracy of the generated queries.


\begin{table*}[!htb]
\centering
\small
\setlength{\tabcolsep}{3pt}
\captionsetup{width=1.0\linewidth}
\caption{\textbf{Case study 1: Performance comparison of different merging methods in japanese Language
processing and mathematical reasoning}. This table presents the performance of the merging methods across a range of Japanese language tasks. The average (Avg) column serves as a comprehensive indicator of each method's overall effectiveness in handling these tasks.}
\label{table:method-comparison}
\begin{tabular}{@{}lrrrrrrrrrrr@{}}
\toprule
& \multicolumn{10}{c}{\textbf{JP Language Model Evaluation Harness}} \\
\cmidrule(l){2-11}
\textbf{Method} & JAQKET & JComQA & JCoLA & XWino & JNLI & MGSM & JSQuAD & MARC & XLSum & Avg &  \\
\midrule
DAM & 0.64 & 0.81 & 0.59 & 0.80 & 0.47 & 0.40 & 0.67 & 0.94 & \textbf{0.22} & \textbf{0.62}  \\
Evolutionary Merging & \textbf{0.71} & 0.80 & 0.57 & \textbf{0.81} & 0.34 & \textbf{0.44} & \textbf{0.76} & \textbf{0.96} & 0.21 & \textbf{0.62}  \\
Model Soups & 0.64 & 0.81 & 0.60 & 0.79 & 0.42 & 0.42 & 0.66 & \textbf{0.96} & \textbf{0.22} & 0.61  \\
DARE-TIES & 0.39 & \textbf{0.82} & \textbf{0.61} & 0.79 & \textbf{0.62} & 0.38 & 0.19 & \textbf{0.96} & 0.20 & 0.55  \\
\bottomrule
\end{tabular}
\end{table*}



\begin{table}[!htb]
\centering
\setlength{\tabcolsep}{3pt}
\captionsetup{width=\linewidth}
\caption{\textbf{Case study 2: Performance comparison of different merging methods in SQL coding, Korean, and German language processing}. This table presents the effectiveness of the merging techniques evaluated on the Okapi (German), KoBEST (Korean), and SQL-Eval benchmarks, along with an overall average (Avg) score across these three benchmarks.}
\label{table:llama-model-comparison}
\resizebox{\linewidth}{!}{
\begin{tabular}{lrrrr}
\toprule
\textbf{Merging method} & German & Korean & SQL & Avg \\
\midrule
DAM & \textbf{0.4434} & 0.5933 & \textbf{0.6125} & \textbf{0.5497} \\
Model Soups & 0.4337 & 0.6121 & 0.5969 & 0.5476 \\
DARE-TIES & 0.4293 & \textbf{0.6307} & 0.5563 & 0.5388 \\
\bottomrule
\end{tabular}
}
\vskip -0.1in
\end{table}

\section{Ablation Studies on the DAM Method}

The overall objective function for DAM, as shown in Equation \ref{objective function}, comprises multiple components, each designed with a specific purpose, as outlined in Section \ref{Objective  Function section}. Here, we analyzed the impact of each component and examined alternative approaches, based on the results in Table \ref{table:ablations}.

\subsection{Comparison of Different Output Distribution Loss Functions}

In place of KL divergence (Equation \ref{KL Div Loss}), we experimented with mean square error (MSE) and entropy loss as potential alternatives for output distribution loss. 

\subsubsection{MSE Loss} 
MSE loss is an intuitive alternative to KL divergence as a loss term to minimize the difference between the output distribution of the merged model and the selected input models. This loss is calculated as shown below:

\[
L_{\text{MSE}} = \frac{1}{N} \sum_{i=1}^{N} \big( \text{logits}_{\text{merged}}(D_i) - \text{logits}_i(D_i) \big)^2
\]

\noindent where \(N\) is the total number of datasets, \(\text{logits}_{\text{merged}}(D_i)\) are the logits of the merged model on dataset \(D_i\), and \(\text{logits}_i(D_i)\) are the logits of the input model \(i\) on dataset \(D_i\).



\subsubsection{Entropy Loss}

Following AdaMerging \cite{yang2023adamerging}, we also considered entropy loss as an alternative for output distribution loss. The objective is to encourage more confident predictions from the merged model. This loss is calculated on the merged model's logits across all datasets:

\[
\mathcal{L}_{\text{entropy}} = \sum_{i=1}^{N} H(\text{logits}_{\text{merged}}(D_i))
\]

\noindent where \(N\) is the total number of datasets, \(H(\cdot)\) is the entropy function, and \(\text{logits}_{\text{merged}}(D_i)\) are the logits of the merged model on dataset \(D_i\).

Minimizing this entropy loss encourages the merged model to make more decisive predictions across all datasets, potentially improving its overall performance and generalization capabilities. Unlike KL divergence or MSE, entropy loss minimizes the entropy of the merged model's output distribution, without reference to the input models




\subsection{Effectiveness of Weight Regularization}
In this experiment, we first evaluated the impact of incorporating cosine similarity loss into the objective function on the performance of the merged model. Specifically, we compared model performance trained using KL divergence or entropy loss functions, both with and without the additional cosine similarity regularization term. 

In addition, we assessed the influence of L1 and L2 regularization term within both the KL-only and KL + Cosine Similarity loss settings to examine their effect on model accuracy and consistency. 

These experiments enabled us to examine how each regularization technique influences the effectiveness and accuracy of the merged model across various configurations, providing insights into their roles in enhancing model performance. 

Throughout the ablation studies, we used the same input models as Case Study 1 and assessed the merged model's performance using the JP Language Evaluation Harness framework. We evaluated the model's average accuracy across the included benchmarks, with each experiment run through three different random seeds to ensure robustness and consistency.



\begin{table}[!htb]
\centering
\footnotesize
\setlength{\tabcolsep}{4pt}
\captionsetup{width=\linewidth}
\caption{\textbf{Ablation studies on the DAM method.} This table presents model evaluation results for various configurations of the DAM objective function, assessed using the JP Language Model Evaluation Harness. Results include the average (Avg) and standard deviation (Std) across three runs with different random seeds.}
\label{table:ablations}
\begin{tabular}{@{}lcc@{}}
\toprule
& \multicolumn{2}{c}{\textbf{JP Language Model}} \\
& \multicolumn{2}{c}{\textbf{Evaluation Harness}} \\
\cmidrule(l){2-3}
\textbf{DAM Objective Function} & Avg & Std \\
\midrule
KL & \textbf{0.62} & 0.034 \\
KL + Reg & 0.60 & - \\
KL + Cosine & \textbf{0.62 }& \textbf{0.007} \\
KL + Cosine + Reg & \textbf{0.62} & \textbf{0.007} \\
MSE + Cosine & 0.53 & - \\
Entropy & 0.60 & 0.019 \\
Entropy + Cosine & 0.60 & 0.027 \\
\bottomrule
\end{tabular}
\end{table}

\section{Results and Discussion}
\label{sec:discussion}

This section presents a detailed analysis of our results, highlighting the comparative performance of DAM, Evolutionary Merging, DARE-TIES, and Model Soups. We discuss insights from the case studies on Japanese and Mathematics domains (Table~\ref{table:method-comparison}) and the German, SQL, and Korean domains (Table~\ref{table:llama-model-comparison}), as well as our findings from ablation studies (Table~\ref{table:ablations}) on DAM’s objective function components.

\subsection{Performance Analysis of DAM Compared to Different Merging Methods}
\subsubsection{Case study 1: Japanese Language Processing and Mathematical Reasoning}

Our experiments in the Japanese and Mathematics domains reveal clear distinctions among merging methods in handling language and reasoning capabilities. 


As shown in Table~\ref{table:method-comparison}, DAM outperforms other methods on average, effectively balancing performance across tasks involving both Japanese comprehension (e.g., \texttt{JAQKET}, \texttt{JComQA}), and mathematical reasoning (\texttt{MGSM}). Evolutionary Merging, while competitive with a 0.62 average score, requires considerable computational resources, highlighting DAM’s advantage in balancing performance with efficiency.

DARE-TIES and Model Soups both show moderate effectiveness, with DARE-TIES achieving higher scores in specific tasks such as \texttt{JComQA} (0.82) and \texttt{MGSM} (0.62). However, their average scores (0.55 and 0.61, respectively) remain below that of DAM, underscoring DAM’s adaptability across diverse tasks without intensive hyperparameter tuning. Model Soups, which is a simple linear averaging approach, performs relatively well, suggesting that linear merging can sometimes rival more complex methods when the merged models are sufficiently similar. Compared to DARE-TIES, Model Soups yields more balanced results across metrics, while DARE-TIES exhibits considerably reduced performance on reading comprehension (\texttt{JSQuAD}) and open-domain question answering (\texttt{JAQKET}) tasks.




\subsubsection{Case Study 2: SQL Coding, German, and Korean Language Processing}

In the case study on German, SQL, and Korean domains, DAM’s effectiveness extends across multilingual and structured data processing tasks, as shown in Table~\ref{table:llama-model-comparison}.


DAM again leads in performance, particularly in German and SQL tasks, with average scores of 0.4434 and 0.6125, respectively. The DAM model performs well across German linguistic tasks with different complexity levels, showing a slight edge over DARE-TIES and Model Soups. In SQL generation tasks, DAM's performance also surpasses Model Soups, which scores 0.5969. DARE-TIES achieves higher effectiveness in Korean tasks (0.6307), likely attributed to differences in model initialization; unlike the other two models, the Korean model was initialized from the Llama 3 base model rather than the Llama 3 Instruct variant. DARE-TIES’s sparsification strategy appears beneficial for minimizing interference in cases where task vectors diverge significantly.

\subsubsection{DAM’s Adaptability in Multilingual Settings}

These observations confirm that DAM’s column-wise scaling effectively balances task-specific representations in multilingual settings. DAM’s flexibility to optimize activation coefficients within individual weight columns proves advantageous in maintaining the distinct linguistic nuances required for each language task.

\subsubsection{Effectiveness of DAM Compared to Evolutionary Merging}

DAM provides a more practical and efficient alternative to Evolutionary Merging. Evolutionary methods require iterative searches through potential parameter configurations, which, while producing competitive performance, demand substantial computational power. In contrast, DAM’s gradient-based optimization achieves similar or superior results with significantly fewer resources. By directly optimizing activation coefficients with a relatively compact dataset, DAM bypasses the extensive computational overhead associated with evolutionary algorithms, enabling faster convergence without compromising on model quality. 

The high performance of DAM’s KL-divergence-based configurations (see Table \ref{table:ablations}) further emphasizes its potential as a scalable alternative to evolutionary techniques. DAM’s architecture-focused merging achieves similar levels of compositional generalization with less manual intervention, making it particularly valuable in environments with limited computational resources.

\subsubsection{Insights on Model Soups and Simplicity in Model Averaging}

One surprising finding is the effectiveness of simple averaging methods like Model Soups. Although it lacks the fine-grained control of DAM or DARE-TIES, Model Soups achieves competitive results in both the Japanese and multilingual domains (e.g., 0.61 average in Table~\ref{table:method-comparison} and 0.5969 for SQL in Table~\ref{table:llama-model-comparison}). This aligns with previous findings on linear interpolation, suggesting that in cases where models share significant similarities, simple averaging may yield satisfactory performance at minimal computational cost. Model Soup’s low resource demand and ease of implementation make it a viable option for scenarios where computational efficiency outweighs the need for nuanced control.

\subsection{Impact of Objective Function Components on DAM’s Performance}

The ablation study (Table~\ref{table:ablations}) reveals the influence of each component in DAM’s objective function on model performance. KL divergence serves as the primary driver, with configurations including KL achieving the highest average scores. The cosine similarity constraint, proposed by ZipLoRA to promote feature diversity among scaling coefficients, was shown to be less effective in this case. Entropy minimization, as demonstrated in AdaMerging’s application, shows potential as a surrogate objective in language modeling tasks, confirming its applicability beyond image classification. While not outperforming KL outright, these results suggest it offers a promising alternative if the domain of the training data is not known. Importantly, the entropy minimization objective also has a lower computational burden, as it does not require logits from the individual input models.

Overall, the combined objective function enables DAM to balance task performance, regularization, and coefficient diversity, leading to a well-integrated merged model suitable for multi-task environments. 


\section{Conclusion}
This paper provides a comprehensive analysis of model merging techniques, spanning from simple averaging methods to automated, data-informed approaches such as evolutionary merging. We introduce DAM as an efficient alternative to evolutionary merging, significantly reducing computational overhead while achieving competitive performance. Our findings challenge the traditional assumption that more complex methods are inherently superior, showing that straightforward techniques like linear averaging can perform just as well, especially when merged models share similar characteristics. Future work could expand these merging strategies to cover a broader range of languages, domains, and modalities, enabling the creation of merged models with effective multi-task capabilities. Additionally, exploring the scalability of these techniques in real-world applications and examining their suitability for resource-constrained environments could further extend their practical impact across the AI landscape.




\bibliography{anthology,custom}
\bibliographystyle{acl_natbib}

\end{document}